\documentclass{article}
\usepackage{ijcai21}

\usepackage{times}

\usepackage{soul}
\usepackage{url}
\usepackage[hidelinks]{hyperref}
\usepackage[utf8]{inputenc}
\usepackage[small]{caption}
\usepackage{graphicx}
\usepackage{color}
\usepackage{amsmath}
\usepackage{booktabs}
\usepackage{tabularx}
\newcolumntype{Y}{>{\centering\arraybackslash}X}
\newcolumntype{Z}{>{\raggedleft\arraybackslash}X}
\usepackage{caption}
\usepackage{xspace}
\usepackage{float}

\newcommand{\system}{{\sc tbc}\xspace}

\newcommand{\remove}[1]{}

\title{Classification of Tabular Data by Text Processing}

\author{Keshav Ramani $^1$
\and
Daniel Borrajo$^{2,3}$
\affiliations
$^1$J.P. Morgan AI Research\\
London, UK\\
$^2$J.P. Morgan AI Research\\
Madrid, Spain\\
$^3$On leave from Universidad Carlos III de Madrid\\
\emails
{keshav.ramani, daniel.borrajo}@jpmchase.com
}

\begin{document}

\maketitle

\begin{abstract}
Natural Language Processing technology has advanced vastly in the past decade. Text processing has been successfully applied to a wide variety of domains. In this paper, we propose a novel framework, Text Based Classification (\system), that uses state of the art text processing techniques to solve classification tasks on tabular data. We provide a set of controlled experiments where we present the benefits of using this approach against other classification methods. Experimental results on several data sets also show that this framework achieves comparable performance to that of several state of the art models in accuracy, precision and recall of predicted classes.

\end{abstract}

\section{Introduction}

Classification tasks are at the core of machine learning. Given some input labeled training instances in some format, the task consists of generating a model (sometimes also called hypothesis) that can take future unlabeled instances and output their label (or class). In connection with this paper, we will focus on two kinds of classification tasks that differentiate on the representation of instances. On the one hand, classical classification tasks have taken as input a table where columns represent features and rows represent instances. A specific column represents the class of each instance. On the other hand, text-based classification tasks in the realm of Natural Language Processing (NLP) usually take a set of labeled documents as input, where each document can be seen as one instance. 

Given the different type of input representation in both cases, different approaches have been followed to solve those tasks. In the case of tabular data, a variety of machine learning techniques have been devised. These techniques use different assumptions, among other factors, on the input representation language (e.g. numerical vs. categorical features), and on the representation bias on the search space~\cite{Mitchell-MLbook}. 

In the case of text-based inputs (documents), the first classification tasks used, and still popular, were solved by converting the input texts into a set of numerical features and then applying standard tabular-based techniques.

Quite often, these numerical features were statistical measures on the frequency of particular words in the document, such as TF-IDF~\cite{jones1972statistical}. More recently, new Deep Learning (DL) techniques have been developed that have completely changed the landscape of classification tasks for NLP~\cite{devlin2018bert}.

In the last two decades, datasets in various fields include text-based features among the other two kinds mentioned before, numerical and categorical. Examples range from applications where only one feature is text to applications where the majority of features are text. For instance, we can mention, among many other applications: recommendation systems that include comments of users as well as other standard features as ratings, or products/services descriptions (both numerical and categorical attributes); chatbots that include conversations with users as well as the corresponding products/services descriptions; click prediction for marketing purposes based on descriptions on previous bought/clicked products;  email analysis where text-based features are mixed up with numerical or categorical attributes; or even recommendation systems for accepting papers at conferences based on the paper's text, reviews, comments, scores on different criteria, and reviewers' metadata.

Given the mixed kinds of inputs, a standard approach would convert the text-based features into a set of numerical features (such as using TF-IDF or related statistical-based transformations), and then apply tabular-based classification techniques. 

In this paper, we propose a new approach to solve tabular-based classification tasks using NLP techniques, that we name Text-Based Classification (\system). Instead of solving NLP tasks transforming them into a tabular representation, we invert the task and solve tabular-based representation classification tasks as document-based classification. Then, we use state of the art DL techniques to solve the original tabular-based classification task. The key of our approach is rooted deeply into Artificial Intelligence: knowledge representation is a fundamental component and automated change of representation allows for an improvement on problem solving in many cases~\cite{Russell95}.

One of the advantages of \system is that it does not require any pre-processing nor hyper-parameter tuning to obtain good results. Careful configuration of these two main components of any learning technique could obtain even better results than the ones we present. However, in this paper, we want to show how a vanilla version of \system can directly outperform other approaches in some tasks.

We first present how our technique works and then we introduce some canonical tasks for which \system can provide benefits over current tabular-based classification techniques. We show the huge difference in performance that can be attained by using our approach in those tasks. Finally, we provide results on some standard datasets to compare the performance of \system with that of other classification techniques. Results show that \system can obtain performance similar to other known techniques in standard classification tasks without any pre-processing or hyper-parameter tuning.

\section{Related Work}

Classification techniques have been widely explored within the realms of Machine Learning~\cite{Mitchell-MLbook}. Some of the most popular classification techniques are tree-based classifiers like Decision Trees~\cite{QuinlanC4.5}, Random Forests~\cite{random-forests} or XGBoost~\cite{chen2016xgboost}. In addition to them, other commonly used classification models include Bayesian Classifiers, neural networks~\cite{RumelhartHintonWilliams86}, or instance-based techniques such as Support Vector Machines~\cite{Cristianini00} or k-nearest neighbours~\cite{duda73}.

On the other hand, the field of Natural Language Processing saw a quantum leap after the advent of Deep Learning (DL). Starting with Recurrent Neural Networks~\cite{rumelhart1985learning} that later were transformed into the Long Short Term Memory Model~\cite{hochreiter1997long}, various sophisticated models were built in order to solve non trivial NLP tasks. Combinations of multiple RNNs and LSTMs have proven to be highly effective and this explains why various state of the art NLP models use them as their building blocks~\cite{devlin2018bert,brown2020language}.

Another dimension along which text processing has improved is that different techniques exist to handle text characters, words and even sentences. Character level LSTMs~\cite{kim2016character} have been around for quite sometime and are widely used for basic sequence related tasks. Sophisticated models like Word2Vec~\cite{mikolov2013efficient}, and Glove~\cite{pennington2014glove} exist at the word level and provide deeper understanding into the meaning of the word and context in which it occurs. Models like Sentence Transformers~\cite{reimers2019sentence} have proven to be effective while working with sentences as well.
DL techniques have been used for NLP in tasks such as text classification~\cite{liu2017deep}, or machine translation~\cite{kalchbrenner2013recurrent,sutskever2014sequence,cho2014properties}.

Recently, some papers have tried to address the task of learning from tabular data using DNNs~\cite{ChengKHSCAACCIA16,abs-1908-07442,Zhang16}. In the first paper~\cite{ChengKHSCAACCIA16}, its authors propose Wide \& Deep learning to integrate a linear regression technique and a deep model to combine their strengths for improving both memorization and generalization in recommendation tasks. The main difference with \system is that they deal with the input features directly, while we convert the input instance into a string and then use only a deep model. In the second paper~\cite{abs-1908-07442}, the authors present TabNet, an approach that does not perform any pre-processing but their focus is on an attention mechanism for feature selection. In the third paper~\cite{Zhang16}, the authors propose different ways of performing embeddings over tabular data. They do not use DNNs as text processing tools as \system does. Further, the work of \cite{DBLP:journals/corr/abs-2106-11189} throws light on the effectiveness of Multi Layer perceptrons and regularization. They study the effects of using a multi layer perceptron with multiple regularization techniques and find that such a technique is likely to outperform classical systems like XGBoost, etc. However, they do not treat this problem as a textual task and instead focus on searching through the phase of different regularization functions. The work of \cite{gorishniy2023revisiting} is similar in spirit and studies the performance of a customized transformer based model and contrast it with Gradient Boosted Decision Trees and find that in some cases the GBDTs outperform deep learning models. This work still does not study the input as a string of text and rather treats numerical and categorical inputs separately by using a custom built tokenizer.

\section{\system}

\begin{figure}[hbt]
    \centering
    \includegraphics[width=\columnwidth, page = 2]{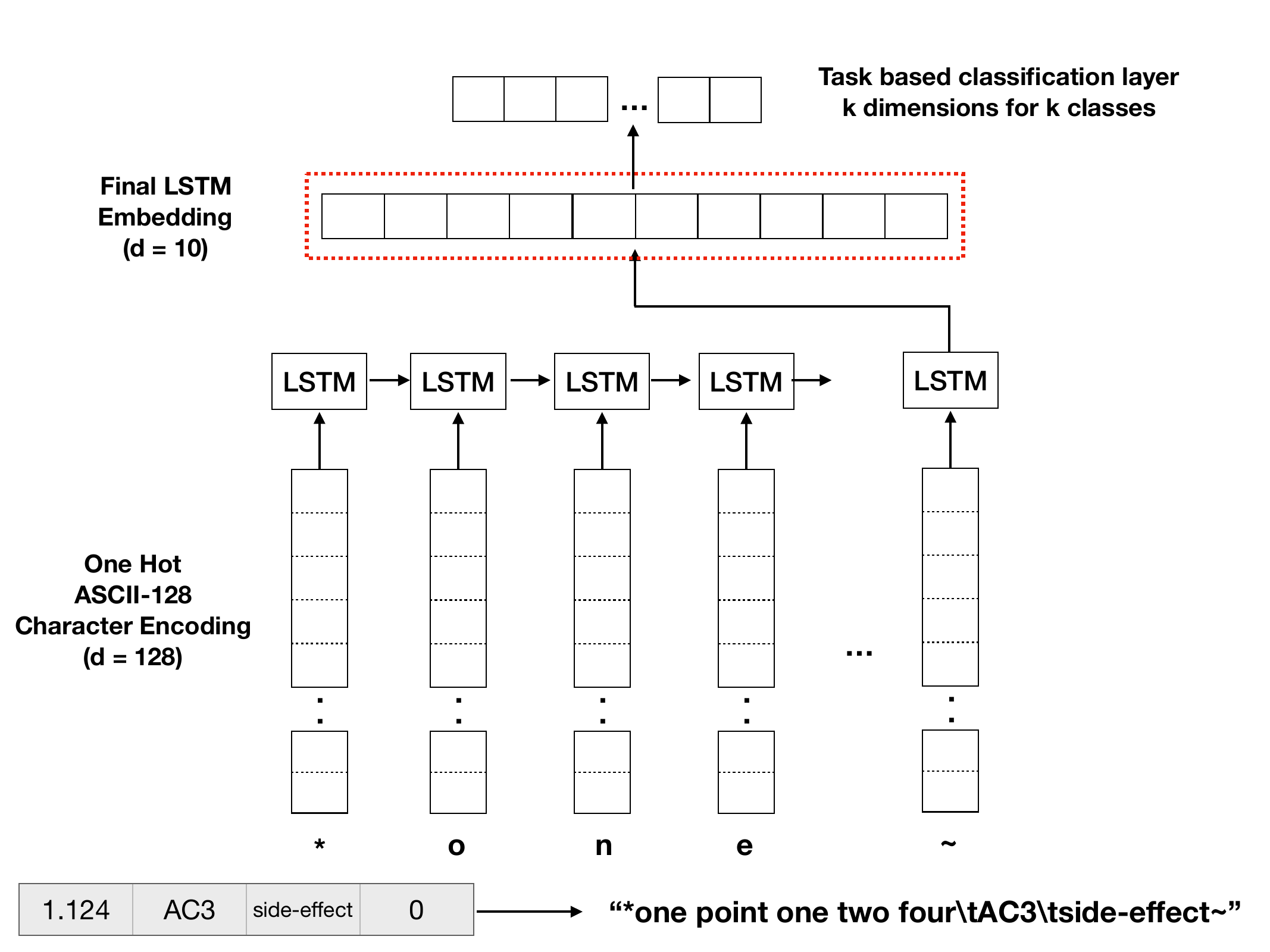}
    
    \caption{Architecture of \system.}
    \label{fig:lstm}
\end{figure}

Figure~\ref{fig:lstm} shows \system's architecture. It takes as input a set of training instances and generates as output a classifier. The main components are two: pre-processing and learning. The pre-processing component translates every feature to its corresponding text, adds some delimiter information and translates the class. The learning component is based on LSTM.

\subsection{Pre-processing data}

Every instance of the input tabular data is ingested by the text processor, and the following transformations are done to generate a string for each instance that:

\begin{enumerate}
    \item Begins with an Asterisk(*)
    \item Categorical and string based attributes for every row are converted to text
    \item Numeric attributes are converted to text by replacing every digit with its textual representation. E.g. 12.3 would be converted to "one two point three".
    \item Each attribute of the same instance is separated by a tab character ('\textbackslash t')
    \item Ends with a Tilde('$\sim$')
\end{enumerate}

The input instance is processed from left to right. Let us consider the following instance as an example. Feature1 is a numeric attribute, Feature2 is categorical, Feature3 is a string and the corresponding class label is 0.

\begin{table}[H]
    \centering
    \begin{tabularx}{\columnwidth}{YYYY}
        \toprule
        Feature1 & Feature2 & Feature3 & Class\\
        \midrule
        1.124 & AC3 & side-effect & 0\\
        \bottomrule
    \end{tabularx}
\end{table}

The output would be the string:

\textbf{*one point one two four\textbackslash tAC3\textbackslash tside-effect$\sim$"}

This training sample is paired with the corresponding class label of 1. In this fashion, we convert all the rows of the input table, pair them with their corresponding class label and generate the data that will be consumed downstream by the text classifier.

\subsection{Learning a classifier}

\system is a Deep Learning based system that makes use of an LSTM at it's heart. The system processes the outputs from the previous module by treating the string as a sequence of characters. Each character is fed into \system as a one-hot encoded vector of 128 dimensions. We choose this because it aligns well with the ASCII-128 model.

\system uses a single LSTM cell followed by a classification layer. Since the use of LSTMs has become common and our implementation is almost identical to the original, we skip an exhaustive description of the LSTM architecture. Our LSTM has 128 inputs, one layer and 10 output states.

We use such a base learner to show how we can achieve good performance even using such a vanilla version of a text processing technique.
This LSTM layer is then directly connected to an output layer with the required number of classes. The model is governed by the Adam optimizer~\cite{kingma2014adam} and the categorical cross entropy is used as the loss function.

Throughout this paper, we train this model for 20 epochs, with a batch size of 1 unless specified.

\section{Controlled Experiments}

In this section, we provide some insights on which classification tasks \system can potentially obtain an improvement in performance over that of some current classification techniques. One might claim that some of the following examples could be successfully solved by current techniques by combining task-dependent pre-processing of input data and/or careful hyper-parameter tuning of the learning techniques. However, as we will show, \system does not need either of these two steps to obtain good performance. We used the input data directly and we did not choose the parameters of the underlying text processing techniques. We can always also tune the parameters of these techniques to even obtain a higher performance. 

We use decision trees for discussion, since they are a key component of widely used techniques for classification (as XGBoost). We also provide comparison against SVMs to showcase the performance of a very different kind of classification technique. 
Another key issue relates to how different classification techniques are implemented in different software packages. We use here the decision tree implementation of the widely used scikit-learn python package. This implementation requires all features to be numeric. While this requirement goes beyond the original Quinlan's implementation of decision trees, it has become a standard in some packages. This forces the developers to perform yet another change of representation in case of using categorical features into a one-hot encoding or equivalent transformation. The analysis we provide is independent of such change of representation and would also apply in case the original representation would had been used.

The tasks we have selected aim at highlighting types of relations among features and the class that will appear when datasets have a mixture of different kinds of feature values, specially when they are categorical or text. We have devised three types of cases: string equivalence, substring matching, and checking properties of numbers. The following subsections deal with each type of case. We first provide the performance of \system and study how it compares against the decision tree and SVM approaches. And then we  analyze the effect of varying the number of distinct inputs on the size of a decision tree classifier. Finally, we prune the said decision tree classifier and observe its effect on accuracy. 

\subsection{String equivalence}

We will start with a simple case, that of checking whether two strings are the same one.
In some tasks, the target hypothesis requires, among other tests on feature values, checking whether two features have the same value. If we take the example of decision tree based learning techniques (including decision rules, random forests, XGBoost and the like), the language that describes the hypothesis does not include such a test (feature i = feature j). Therefore, it is well known that the vanilla versions would generate a tree that grows with the number of different values of those features. Unless pruning is used, decision trees are known to overfit and this type of classification problem is an example. This means that they do not actually learn the actual relation among the attributes, but some (or all) specific cases of the relation between the two.

If we add pruning, the size of the trees will naturally decrease, but accuracy will also decrease since many of those combinations of pairs of values will be considered as noisy observations in some tree leaves.  Using ensembles of trees, as random forests, we would randomly cover a subset combinations (or all with a big enough number of trees), so we would be able to recover accuracy at the cost of increasing the size of the model again. Section~\ref{sec:size-accuracy} presents some analysis on this trade-off.

The training set for this task was constructed with 1000 rows and 2 attributes. Each attribute contains 1000 randomly generated words.
Now, some rows were selected at random ($p=0.5$), and the value of the first attribute was assigned to the second. As a result, the training set now has 487 rows with matching attributes and 513 rows that do not. Similarly, a test set of 500 rows was  generated - with 241 rows where both attributes matched and 259 rows where they didn't. The Decision Tree and Support Vector Machine that are being compared with, use the default parameters from the sklearn package. For both these models, we one-hot encode the string values on both columns and pass them to the model. For \system, we merely apply the described pre-processing.

Table~\ref{tab:1} shows the results of comparing \system's performance in terms of precision, recall and accuracy to that of decision trees (DT) and SVM. They are measured using either the training or test sets. We also show the total training and test time. \system  outperforms them on the test set in precision and accuracy. Its recall is very close to that of the best technique, decision trees. As we can see, our technique suffers less from overfitting, as suggested by the difference between the accuracy in training and test. We can also see that precision significantly drops in the case of decision trees and SVMs. 

\begin{table}[hbt]
    \centering
    \begin{tabularx}{\columnwidth}{XYZZZZ}
        \toprule
        Model & Set & Precision & Recall & Accuracy & Time\\ \midrule
        DT & Train & 1.00 & 1.00 & 1.00& 1.59s\\
        & Test & 0.08 & \textbf{0.66} & 0.54&0.02s\\
        \midrule
        \system & Train & 0.91 & 0.87 & 0.89&158s\\
        & Test & \textbf{0.69} & 0.64 & \textbf{0.67}&0.34s\\
        \midrule
        SVM & Train & 0.99 & 1.00 & 0.99&3.60s\\
        & Test & 0.07 & 0.56 & 0.52&1.79s\\
        \bottomrule
    \end{tabularx}
    \caption{String Equivalence - Performance of different models.}
    \label{tab:1}
\end{table}

\subsection{Substring matching}

A generalization of the previous task consists of finding text-based relations among attributes. Examples are when one feature is a substring of another one, a synonym or a translation in another language of another feature. In all those cases, traditional learning techniques will not be able to successfully find the corresponding relation, while \system leverages on the use of text processing to find those relations and how they help the classification task.

Let us focus on substring matching. The class is 1 if one feature value is a substring of another feature value. 
The training and test sets were generated using the same procedure as the previous experiment. Table~\ref{tab:2} shows the result of this task. We can see similar trends as in the previous experiment. Decision trees performance hugely decreases in unseen instances. \system accuracy also decreases, but it still obtains the highest accuracy over the three compared systems. SVMs obtain the best recall, but at the cost of significantly decreasing their precision.

\begin{table}[hbt]
    \centering
    \begin{tabularx}{\columnwidth}{XYZZZZ}
        \toprule
        Model & Set & Precision & Recall & Accuracy & Time\\ 
        \midrule 
        DT & Train & 1.00 & 1.00 & 1.00& 0.92s\\
        & Test & 0.45 & 0.80 & 0.68&0.02s\\
        \midrule
        \system & Train & 0.88 & 0.80 & 0.83&159s\\
        & Test & \textbf{0.73} & 0.69 & \textbf{0.71}&0.33s\\
        \midrule
        SVM & Train & 0.99 & 1.00 & 0.99&3.27s\\
        & Test & 0.38 & \textbf{0.82} & 0.66&1.62s\\
        \bottomrule
    \end{tabularx}
    \caption{Substring Matching - Performance of different models.}
    \label{tab:2}
\end{table}

\subsection{Checking number properties}

In some tasks, some features might be numeric. \system would not behave well if the relation among features requires some kind of mathematical relation among them. However, it can behave better than other techniques when the right classification rule has to check some property of those numbers. For instance, whether they are odd, start with a given digit, or contain a given digit in the number. Given that we convert numbers into their corresponding text representation, all these tasks can be successfully addressed by \system, while they would become really hard for other learning techniques.

Let us focus on the task of detecting an odd number. The dataset for this task was constructed with one feature and 1000 rows. The value of the features was filled by uniformly randomly sampling floating point numbers between 0 and 9,999. 800 rows were randomly sampled to constitute the training set and 200 made up the test set. The training set has 401 instances of odd numbers, and 399 with even ones. The test set has 93/107 instances of each class, respectively.

Table~\ref{tab:3} shows the results of the experiment. We see the huge difference of using \system over the other two approaches. We can see that the values of those metrics measured over training instances is perfect. However, the performance decreases on unseen instances for decision trees, while it is still perfect for \system. In the test set, we included unseen values for those features and that explains why the performance of decision trees decreases. Instead, \system is not affected by unseen examples and is able to generalize well by using a language model of instances.

\begin{table}[hbt]
    \centering
    \begin{tabularx}{\columnwidth}{XYZZZZ}
        \toprule
        Model & Set & Precision & Recall & Accuracy & Time\\ 
        \midrule 
        DT & Train & 1.00 & 1.00 & 1.00 & 0.004s\\
        & Test & 0.42 & 0.42 & 0.38&0.001s\\ \midrule
        \system & Train & 1.00 & 1.00 & 1.00&130s\\
        & Test & \textbf{1.00} & \textbf{1.00} & \textbf{1.00}&0.288s\\ \midrule
        SVM & Train & 0.37 & 0.53 & 0.52&0.020s\\
        & Test & 0.21 & 0.38 & 0.39&0.004s\\
        \bottomrule
    \end{tabularx}
    \caption{Number properties - Performance of different models.}
    \label{tab:3}
\end{table}

\subsection{Size and accuracy}
\label{sec:size-accuracy}

\begin{figure*}[hbtp]
    \centering
    \includegraphics[width=12.5cm, height = 6cm, page = 1]{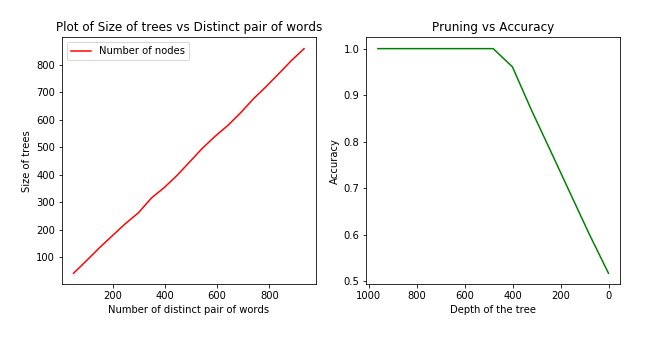}
    
    (a) String equivalence task.
    
    \includegraphics[width=12.5cm, height = 6cm, page = 1]{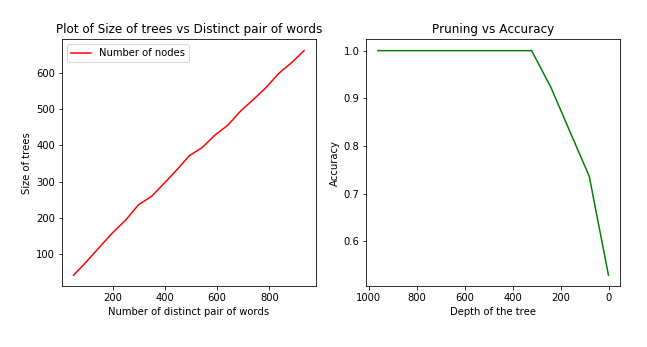}
    
    (b) Substring matching task.
    
    \includegraphics[width=12.5cm, height = 6cm, page = 1]{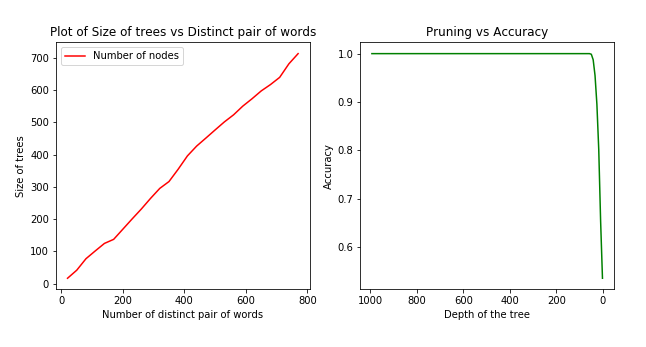}
    
    (c) Checking number properties task.
    
    \caption{Size of trees as a function of the number of different combinations in the instances and train accuracy as a function of pruning. It shows results in the (a) string equivalence, (b) substring matching and (c) checking number properties tasks.}
    \label{fig:size-accuracy}
\end{figure*}

Figure~\ref{fig:size-accuracy} analyzes size and accuracy of decision trees. On the left, we show how the number of different combinations of values of the features affect the size of the tree. As we can see, the size increases linearly with the number of different combinations of values of those features. 


\begin{table*}[hbt]
   \centering
   \begin{tabularx}{\textwidth}{YZZZZZZZZZZZZZZZZ} 
       \toprule
         & & \multicolumn{3}{c}{XGBoost} & \multicolumn{3}{c}{\system} & \multicolumn{3}{c}{Random Forest} & \multicolumn{3}{c}{SVM} & \multicolumn{3}{c}{ANN}\\
       \cmidrule(lr){3-5}
       \cmidrule(lr){6-8}
       \cmidrule(lr){9-11}
       \cmidrule(lr){12-14}
       \cmidrule(lr){15-17}
       Dataset & Set & A & P & R & A & P & R & A & P & R & A & P & R & A & P & R \\
       \midrule
       Adult & Train & 90	& 84	& 72	& 84 & 70 & 59 & 99 & 100 & 99 & 86 & 78 & 61 & 89 & 84 & 68\\
       & Test & 87	& 77	& 65	& 84 & 70 & 59 & 85 & 72 & 62 & 85 & 75 & 59 & 84 & 74 & 58\\
       \midrule
       Titanic & Train & 96 & 98 & 93 & 80 & 81 & 64 & 98 & 99 & 96 & 83 & 83 & 73 & 81 & 80 & 69\\
       & Test & 81 & 77 & 72 & 78 & 76 & 63 & 80 & 74 & 73 & 83 & 82 & 72 & 86 & 83 & 77\\
       \midrule
       Iris & Train & 100	& 100	& 100	& 87	& 87	& 87 & 100 & 100 & 100 & 97 & 97 & 97 & 67 & 67 & 67\\
       & Test & 96	& 96	& 96	& 55	& 55	& 55 & 96 & 96 & 96 & 96 & 96 & 96 & 80 & 80 & 80\\
       \midrule
       Dress & Train & 99	& 99	& 99	& 66	& 62	& 49 & 99 & 99 & 99 & 81 & 93 & 61 & 98 & 98 & 97\\
       & Test & 57	& 50	& 44	& 54	& 40	& 26 & 58 & 48 & 36 & 60 & 53 & 30 & 58 & 48 & 56\\
       \bottomrule
   \end{tabularx}
   \captionsetup{justification=centering}
   \caption{Experiments on public datasets. A stands for accuracy, P for precision and R for recall.}
   \label{tab:4}
\end{table*}

On the right, we show how the accuracy on the training set is affected by different levels of pruning (setting the max depth of the tree). We see an abrupt decrease in accuracy when we increase the pruning (less max depth). The change in accuracy is noticeable earlier in the first task (starting with depths of the tree - pruning - around 400). In case of pruning at depth of 200, the accuracy already decreases to 0.7 from 1.0. In the last experiment, accuracy only drops with strong pruning (shallow max tree depth).

\section{Experiments}

In order to understand how \system works with multi-type data, we compared its performance against other representative techniques: XGBoost, Random Forest, SVMs and ANN. We used four standard datasets: Adult \cite{misc_adult_2}, Titanic \cite{OpenML2013}, Iris \cite{misc_iris_53} and Dress \cite{misc_dresses_attribute_sales_289}. The default parameters were used for the XGBoost models, Random Forest and SVMs. We constructed a two-layer NN with twice as many neurons as the input dimensions in the first layer, followed by a softmax classification layer. The activation of the first layer was ReLU. We used the Adam optimizer and the Categorical Cross Entropy as the loss function. A batch size of 32 was for all the experiments in this section. \system had to use a batch size of 1 for the Iris and Dress dataset due to the smaller sizes of these datasets.

For every task and model, we report the Accuracy, Precision and Recall. In the case of Iris alone, we report the micro average Precision and Recall. This is because this dataset is not used for a binary classification task, but for a 3 way classification task instead. The scores for all models have been reported based off 5-fold cross validation.

These datasets are not tailored towards text processing, so there should not be any advantage of \system over the other compared techniques. Even so, we wanted to show that the performance of \system is not far from other techniques.  As we discussed in the previous section and in the introduction, the more adequate datasets for \system would be those where there are one or more text features, combined with other kinds of features. However, most datasets that are available that contain text-related features lie in two categories: they either present a single attribute with the text as a document (e.g. Reuters); or they have carried out a pre-processing step of computing TF-IDF over the documents converting the text attributes into real-valued attributes. None of these two variants are the best option for \system. In case there is only one text as a single attribute, using \system would be equivalent to using standard LSTM-based classifiers. In the second case, we would benefit of dealing directly with the input text, but it is not available. Also, the comparison against other non-text based techniques requires a transformation step of text for other techniques to convert text features into numeric features (TF-IDF or equivalent).

Table~\ref{tab:4} shows the results. A stands for accuracy, P for precision and R for recall. We can see that the accuracy measured with cross-validation is close to that of the other techniques for all datasets except on the Iris dataset. This dataset only includes numeric attributes and the class relates to numerical relations between attributes and the class. This numerical relations can hardly be represented in \system. For instance, a decision tree with a high accuracy will contain tests such as 'petalwidth $<=$ 0.6'.

\section{Conclusion and future work}

We have presented \system, an approach to solve classification tasks by using NLP techniques. The advantage of \system over other techniques is that it can benefit from text-based analysis of feature values and the mutual relations among feature values and the class, so that it can naturally deal with data that includes text-based features. Also, we have shown that it already has good performance without any kind of hyper-parameter tuning or pre-processing of the input data. 

The experimental results show that it can achieve  equivalent performance to that of other techniques with a vanilla version of an LSTM in datasets that do not have the best characteristics for our technique.

In future work, we would like to include DL models that perform word embeddings in order to deal with even richer classification tasks where the meaning of different feature values and class is also relevant. 

\section{Disclaimer}
This paper was prepared for informational purposes by the Artificial Intelligence Research group of JPMorgan Chase \& Co and its affiliates (“J.P. Morgan”) and is not a product of the Research Department of J.P. Morgan.  J.P. Morgan makes no representation and warranty whatsoever and disclaims all liability, for the completeness, accuracy or reliability of the information contained herein.  This document is not intended as investment research or investment advice, or a recommendation, offer or solicitation for the purchase or sale of any security, financial instrument, financial product or service, or to be used in any way for evaluating the merits of participating in any transaction, and shall not constitute a solicitation under any jurisdiction or to any person, if such solicitation under such jurisdiction or to such person would be unlawful.   

© 2023 JPMorgan Chase \& Co. All rights reserved 

\bibliographystyle{named}
\bibliography{main.bib}

\end{document}